%% file: root.tex
\documentclass[letterpaper, 10 pt, conference]{ieeeconf}  

\IEEEoverridecommandlockouts                              

\usepackage{cite}
\usepackage{amsmath,amssymb,amsfonts}
\usepackage{algorithmic}
\usepackage{graphicx}
\usepackage{textcomp}
\usepackage{xcolor}
\usepackage[nolist]{acronym}
\usepackage{subcaption}
\usepackage{booktabs}
\usepackage{comment}
\usepackage{hyperref}
\usepackage{tikz}

\acrodef{krr}[KR\&R]{knowledge representation and reasoning}
\acrodef{llm}[LLM]{large language model}
\acrodef{ai}[AI]{artificial intelligence}
\acrodef{nlp}[NLP]{natural language processing}
\acrodef{tamp}[TAMP]{Task and Motion Planning}
\acrodef{rl}[RL]{reinforcement learning}
\acrodef{mae}[MAE]{Mean Absolute Error}
\acrodef{rmse}[RMSE]{Root Mean Square Error}
\acrodef{max}[MAX]{Maximum Absolute Error}
\acrodef{knn}[k-NN]{k-Nearest Neighbors}
\acrodef{rdf}[RDF]{Resource Description Format}
\acrodef{owl}[OWL]{Web Ontology Language}
\acrodef{pcl}[PCL]{Point Cloud Library}
\acrodef{vr}[VR]{virtual reality}

\newcommand\copyrighttext{%
  \footnotesize \textcopyright 2024 IEEE. Personal use of this material is permitted.
  Permission from IEEE must be obtained for all other uses, in any current or future 
  media, including reprinting/republishing this material for advertising or promotional 
  purposes, creating new collective works, for resale or redistribution to servers or 
  lists, or reuse of any copyrighted component of this work in other works.}
\newcommand\copyrightnotice{%
\begin{tikzpicture}[remember picture,overlay]
\node[anchor=south,yshift=10pt] at (current page.south) {\fbox{\parbox{\dimexpr\textwidth-\fboxsep-\fboxrule\relax}{\copyrighttext}}};
\end{tikzpicture}%
}

\title{\LARGE \bf
RoboGrind: Intuitive and Interactive Surface Treatment with Industrial Robots
}

\author{Benjamin Alt$^{1,2,*}$, Florian Stöckl$^{3}$, Silvan Müller$^{3}$,  Christopher Braun$^{4,5}$, Julian Raible$^{4,5}$,\\Saad Alhasan$^{3}$, Oliver Rettig$^{3}$, Lukas Ringle$^{1}$, Darko Katic$^{1}$, Rainer Jäkel$^{1}$,\\Michael Beetz$^{2}$, Marcus Strand$^{3}$, and Marco F. Huber$^{4,5}$
\thanks{$^{1}$ArtiMinds Robotics, 76131 Karlsruhe, Germany}
\thanks{$^{2}$Institute for Artificial Intelligence, University of Bremen, 28359 Bremen, Germany}
\thanks{$^{3}$Baden-W\"urttemberg Cooperative State University, 76133 Karlsruhe, Germany}%
\thanks{$^{4}$Institute of Industrial Manufacturing and Management IFF, University of Stuttgart, 70569 Stuttgart, Germany}%
\thanks{$^{5}$Fraunhofer Institute for Manufacturing Engineering and Automation IPA, 70569 Stuttgart, Germany}%
\thanks{$^{*}$ Corresponding author: {\tt\small benjamin.alt@uni-bremen.de}}
}

\begin{document}

\maketitle
\copyrightnotice
\thispagestyle{empty}
\pagestyle{empty}

\begin{abstract}

Surface treatment tasks such as grinding, sanding or polishing are a vital step of the value chain in many industries, but are notoriously challenging to automate. We present RoboGrind, an integrated system for the intuitive, interactive automation of surface treatment tasks with industrial robots. It combines a sophisticated 3D perception pipeline for surface scanning and automatic defect identification, an interactive voice-controlled wizard system for the AI-assisted bootstrapping and parameterization of robot programs, and an automatic planning and execution pipeline for force-controlled robotic surface treatment. RoboGrind is evaluated both under laboratory and real-world conditions in the context of refabricating fiberglass wind turbine blades.

\end{abstract}

\section{Introduction}
In a wide range of industries such as aerospace, consumer goods manufacturing, or the energy sector, surface treatment tasks are integral components of the value chain. With the rise of remanufacturing as a core component of the circular economy, the need for robust, cost-efficient \mbox{(re-)finishing} of surfaces has become even more pressing. One example is the refabrication of wind turbine rotor blades, which require considerable surface treatment after several years of use. Due to a shortage of qualified labor and high costs, economically feasible remanufacturing requires novel, robust solutions for automating surface finishing tasks with robots. 

Robotic surface treatment is challenging due to the physics of contact and abrasion, which are hard to simulate and require robust force control. Moreover, in the context of remanufacturing, workpieces have been subject to different degrees of wear, requiring sophisticated perception and algorithms for automatically identifying defects. At the same time, many steps in the surface treatment workflow, from the proper parameterization of robot programs to the choice of tool, require considerable human expertise.
We hypothesize that a software system that combines sophisticated perception, planning, reasoning, and control capabilities can greatly reduce the cost of automation for surface finishing tasks. We also propose that the human is an integral source of knowledge, guidance, and oversight, and ought to remain involved in the surface finishing process, though in a different capacity and to a lesser degree.

\begin{figure}[t]
    \centering
    \includegraphics[width=\linewidth]{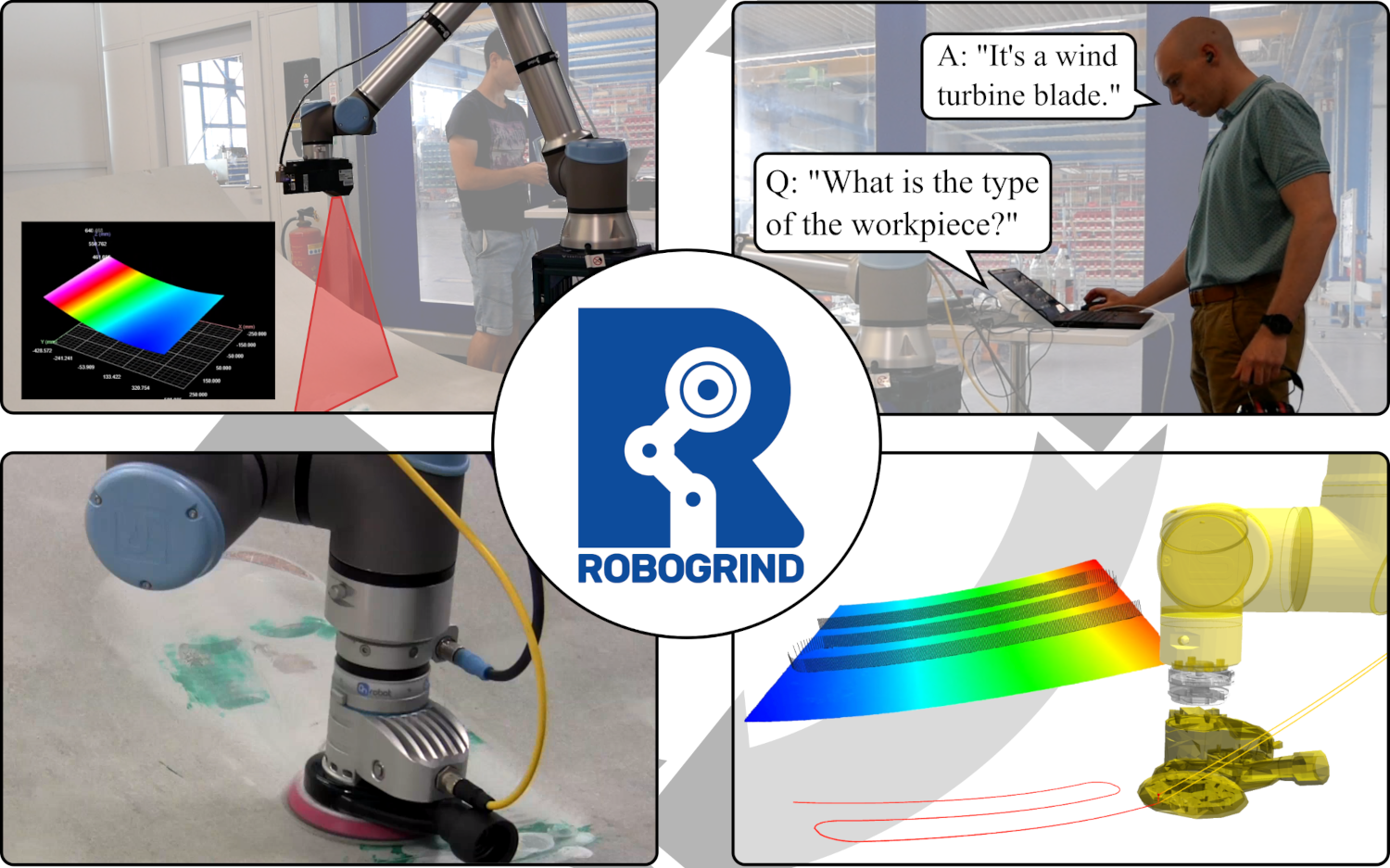}
    \caption{RoboGrind is an intuitive, interactive system for robotic surface treatment comprising perception, program generation, planning and control.}
    \label{fig:teaser-image}
\end{figure}

In this paper, we present RoboGrind, a software system for iteractive,  \ac{ai}-assisted surface treatment with industrial robots. It combines four distinct technical contributions:
\begin{enumerate}
    \item Advanced 3D vision and automatic defect detection, enabling the automatic treatment of different surface geometries and defect locations;
    \item Automatic planning of tool paths based on a 3D scan of the workpiece, depending on the tool and material;
    \item \ac{ai}-assisted bootstrapping of robot programs, with a deep-learning-based \ac{nlp} frontend for user interaction;
    \item Automatic simulation and force-controlled execution of the generated robot programs.
\end{enumerate}
RoboGrind integrates these components into a unified architecture to achieve a very high degree of automation. The perception, planning, and control systems are individually evaluated under laboratory conditions. The overall system is evaluated under real-world conditions for the refabrication of fiberglass wind turbine blades. This paper conducts one of the first investigations into force-controlled disk sanding of fiberglass with lightweight collaborative robot arms.

\section{Related Work}
\subsection{Robotic Surface Treatment}
Li et al. implemented a pipeline for automated rail grinding \cite{Li2018}, which removes excess material from welds. They designed a system with seven main modules, including modules for measurement, motion control, and information feedback. The main difference to our system is their specialization on weld seams and lack of guided adaption of sanding parameters based on a knowledge base. Oyekan et al. researched methods to automate fan-blade reconditioning of aerospace maintenance \cite{Oyekan2020}, using a simulation environment based on a digital twin. Our work ties back to the point they make in their conclusion: The main bottleneck for automated grinding remains ``embedding the knowledge of a skill worker into an automated cell'' \cite{Oyekan2020}.

\subsection{AI-Assisted Robot Programming}
A wide variety of methods facilitating the creation of robot programs have been proposed. \Ac{tamp} approaches view robot programming as a joint task- and motion-level planning problem and combine search-based planning in task space with collision-free motion planning \cite{garrett_integrated_2020, kaelbling_hierarchical_2011}. CRAM \cite{beetz_cram_2010} proposes a knowledge-based approach to robot programming by combining a symbolic program and knowledge representation with the KnowRob \ac{krr} engine \cite{beetz_know_2018} and a realistic simulation environment. This paradigm has been shown to support several forms of \ac{ai}-assisted robot programming \cite{koralewski_self-specialization_2019}, or synthesis of robot programs from human \ac{vr} demonstrations \cite{alt_knowledge-driven_2023}. Several recent approaches propose to leverage natural language as a suitable abstraction for program synthesis. Code as Policies \cite{liang_code_2023-1} and ProgPrompt \cite{singh_progprompt_2022}  use a \ac{llm} to generate Python code that uses an API of robot primitives to solve manipulation tasks. TidyBot \cite{wu_tidybot_2023} uses natural language as an intuitive representation for user input. We propose to combine the advantages of natural language-based systems with structured \ac{krr} and planning to achieve intuitive user interaction and precise, industrially robust execution.

\begin{figure*}[h]
    \centering
    \includegraphics[width=\textwidth]{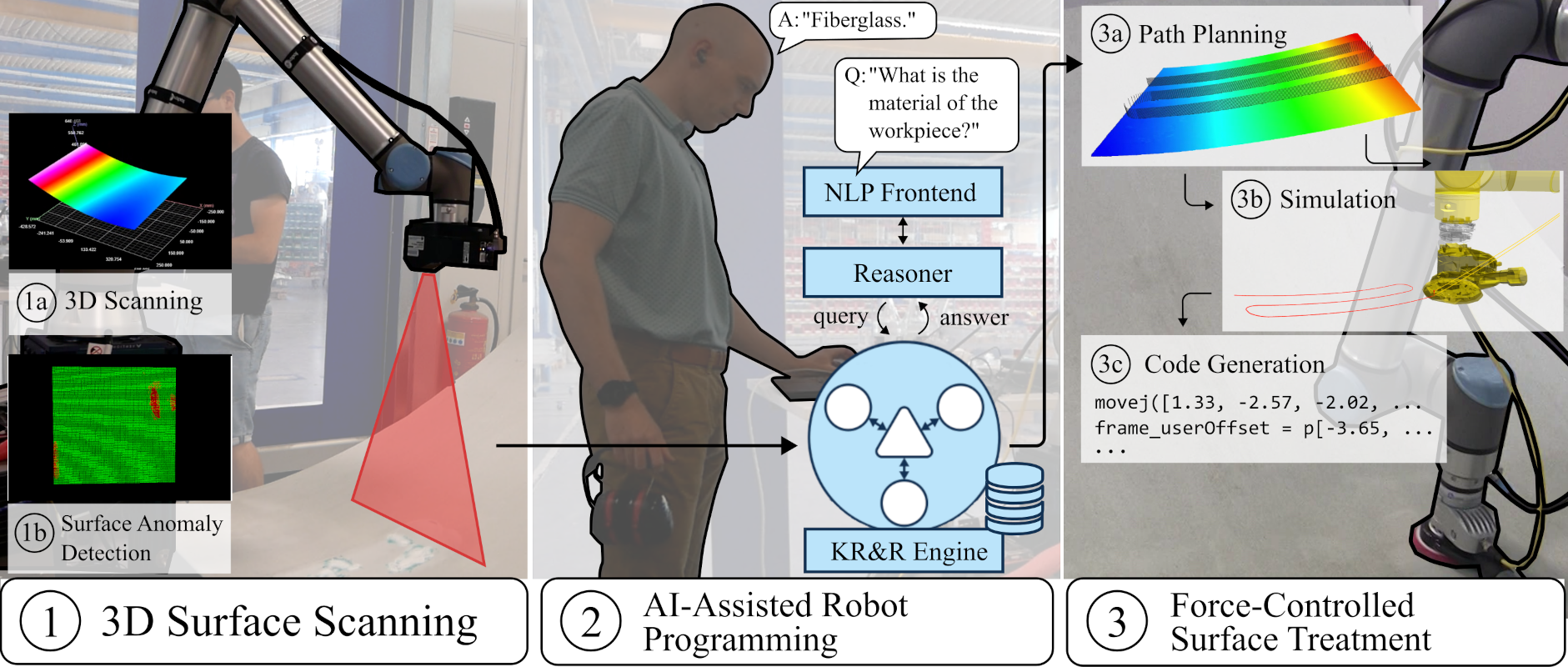}
    \caption{RoboGrind integrates perception (1), AI-assisted programming (2), planning and force control (3) into a comprehensive assistance system for robotic surface treatment.}
    \label{fig:overview}
\end{figure*}

\section{Overview}
RoboGrind is an interactive system for rapidly and intuitively automating surface finishing tasks with robots. It achieves this by combining advanced perception and planning algorithms with \ac{krr} capabilities and natural language understanding. RoboGrind realizes a three-step workflow (see Fig. \ref{fig:overview}):

\subsubsection{3D Surface Scanning}
The workpiece is scanned with a laser scanner. Scanning and scan data processing is nearly completely automated and defects on the workpiece are automatically detected.
\subsubsection{AI-Assisted Robot Programming}
Via natural-language interaction, the user is guided through the process of generating a parameterized robot program for the given task (e.g., sanding, polishing, or deburring). A digital twin of the robot and the required robot skills to perform the task are automatically instantiated and parameterized. A path planner generates a Cartesian tool path depending on the surface geometry.
\subsubsection{Force-Controlled Surface Treatment}
The generated robot program is simulated and displayed to the user for validation. It is then automatically compiled into executable robot code and executed on the robot. This typically involves force-controlled, abrasive motions in contact with the surface, as well as collision-free approach and depart motions. As most surface treatment tasks are iterative processes, the process is iterated until the desired surface quality is reached.

The three steps of the workflow are detailed in the following sections.

\section{3D Surface Scanning}
\label{sec:perception}
RoboGrind performs program generation and path planning based on a 3D point cloud representation of the surface. We use a Gocator 2490\footnote{\href{https://lmi3d.com/}{LMI Technologies GmbH}, Teltow, Germany} laser line scanner with a field of view of up to 2\,m and a resolution of 6\,$\mu$m. The starting point of a scan is defined manually according to the measurement range. Segmentation, outlier detection, and downsampling are performed. Based on an initial calibration step, the point cloud is transformed into the robot's base frame. We propose an automatic defect detection pipeline that combines two outlier detection algorithms: First, curve fitting by linear regression is performed separately for the points of each scanned line and, based on several thresholds, outliers and defects are marked and excluded. Second, a StatisticalOutlierRemoval filter from the \acf{pcl} based on \ac{knn} is used to detect points with few neighbors. This allows detecting scanning artefacts that cause holes in point clouds, and eliminates outliers that do not belong to a scanned part. It also helps reducing the number of false positives detected, as these have no high point density and have too few neighbors. The combined algorithms can detect all defects that deviate from an ideal surface shape, including dents, bumps, scratches, and rough surface areas. Unlike neural networks, they require no training. The code can be found in our GitLab repository\footnote{https://gitlab.com/rahm-lab/robogrind-surfacetreatment}.

\section{AI-Assisted Robot Programming}
\label{sec:ai-assisted-robot-programming}
A core objective of RoboGrind is to solve robot-based surface treatment tasks with a high degree of automation. RoboGrind builds upon a \ac{krr} engine to automate large parts of the surface treatment workflow. At the same time, the complexity of the task as well as safety considerations necessitate the continued involvement of human experts in the process. Recognizing this, RoboGrind realizes a robot programming paradigm that is fundamentally interactive, and integrates a natural language interface to facilitate intuitive interaction with the user.
\subsection{Meta-Wizard Architecture}
The cognitive subsystem for interactive robot programming, which we call \emph{meta-wizard}, comprises a knowledge base and a novel reasoner for surface treatment tasks. Both are integrated into the KnowRob \ac{krr} engine \cite{beetz_know_2018}, which provides a framework for robot cognition and metaprogramming.
\subsubsection{Knowledge Base}
The knowledge base contains commonsense, robot-specific and domain-specific knowledge and forms the basis for reasoning. In the KnowRob framework, knowledge is stored as \ac{rdf} triples in a MongoDB database. KnowRob provides utilities for the automatic conversion of \ac{owl} ontologies into \ac{rdf} triples, which in turn allows reasoners to jointly reason over symbolic (ontological) knowledge and subsymbolic, unstructured data such as force-torque data from a robot \cite{beetz_know_2018}. Commonsense and robot-specific knowledge is provided via the SOMA upper-level ontology \cite{besler_foundations_2020}, while additional domain-specific knowledge is represented in a novel RoboGrind ontology. This ontology extends SOMA by surface treatment-specific concepts such as task types (e.g., \texttt{Sanding}), tools (e.g., \texttt{OrbitalSander}), or materials (e.g., \texttt{Fiberglass}) as well as their subsymbolic properties (e.g., rotational speeds). The knowledge base can be extended to new domains, materials, or tools by adding new classes and individuals to the ontology.\\
\begin{figure*}
    \centering
    \includegraphics[width=\textwidth]{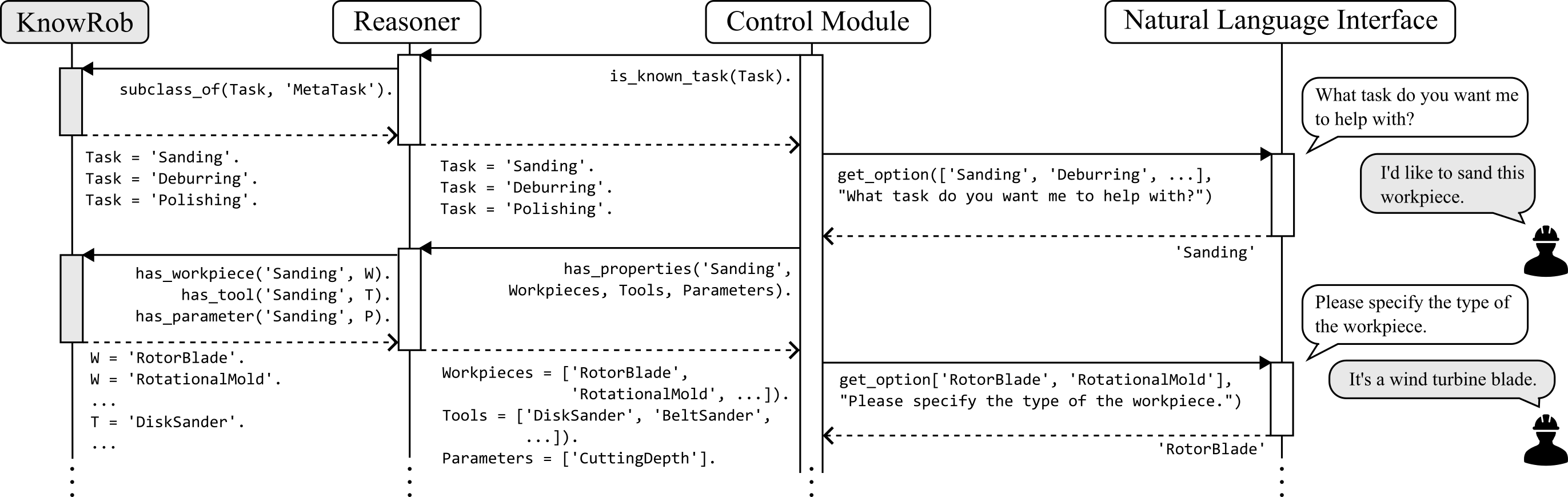}
    \caption{Illustration of the meta-wizard's interactive symbol grounding mechanism. The meta-wizard's main control module retrieves symbolic knowledge about tasks, workpieces, etc., via Prolog queries to a domain-specific reasoner connected to the KnowRob \ac{krr} engine. Missing information about the concrete task, workpiece, etc., is obtained via dialog with the user.}
    \label{fig:meta-wizard-sequence-diagram}
\end{figure*}
One important feature of the meta-wizard subsystem is that the surface treatment workflow itself is represented as classes and individuals of the ontology. Using \texttt{soma:Workflow} and related classes, surface treatment processes can be represented by modeling their respective steps (e.g., \texttt{SurfaceScanning}, \texttt{Simulation}, \texttt{Execution}, \texttt{QualityControl}) and chaining them via succeedence relations. This representation of workflows as ontological knowledge permits reasoners to reason about the workflow itself, and enables the meta-wizard to change its behavior depending on its belief state---its ontological knowledge and current understanding of the task and environment. Moreover, it permits human experts to add procedural knowledge to the wizard without changing its code.

\begin{figure}
\centering
\begin{minipage}[b][][b]{.57\linewidth}
      \centering
    {\includegraphics[width=0.95\linewidth]{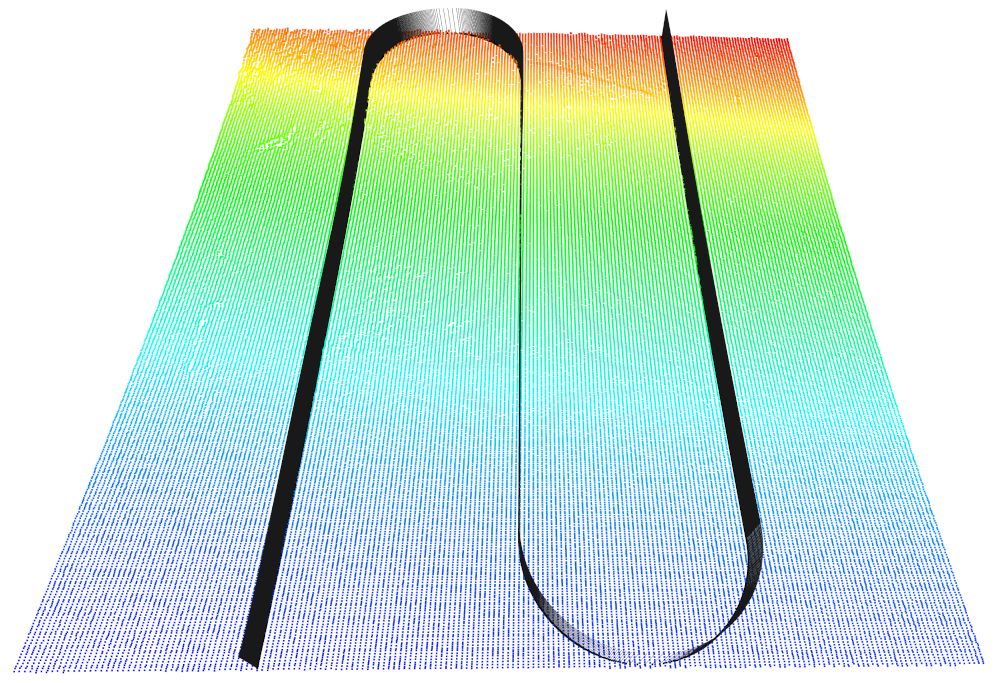}}
    \captionof{figure}{Path planning on point cloud data.}
    \label{fig:path_planning}
\end{minipage}%
\hfill%
\begin{minipage}[b][][b]{.42\linewidth}
\centering
  \includegraphics[width=\linewidth]{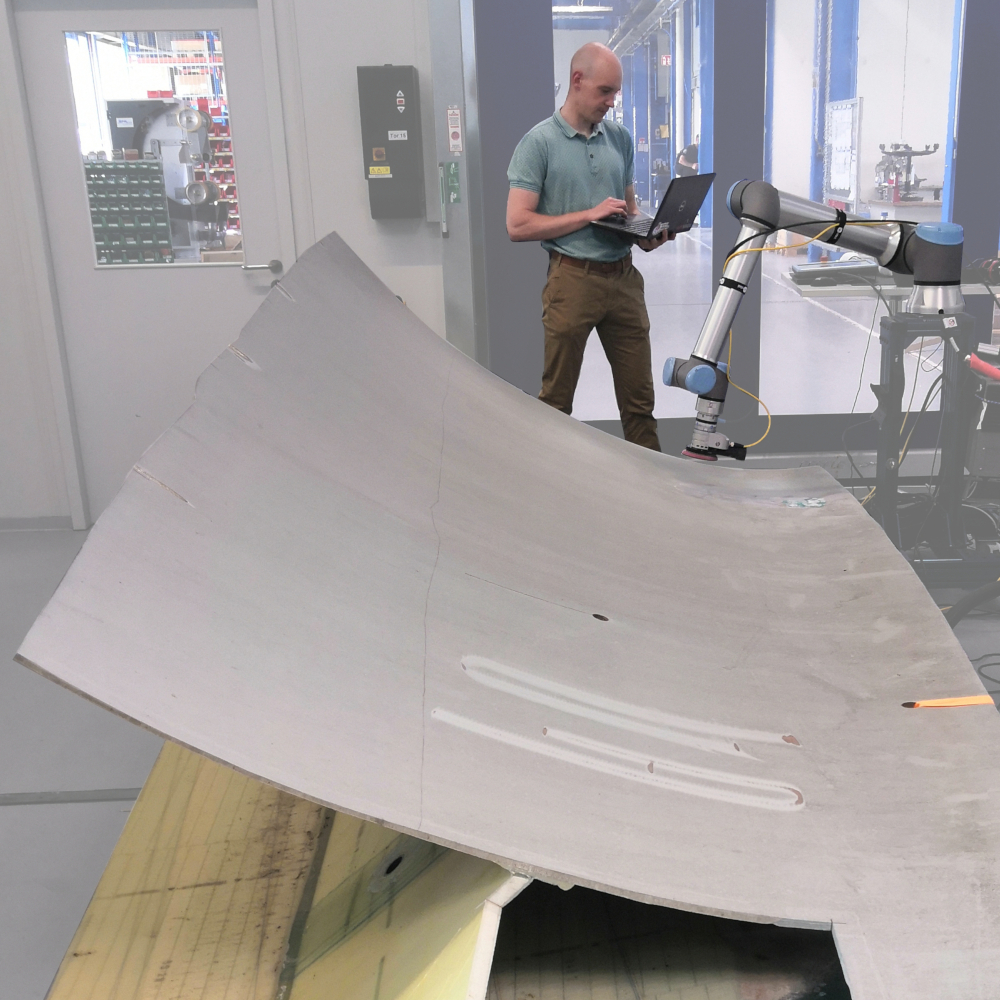}
  \captionof{figure}{Test setup under real-world conditions.}
  \label{fig:shl-robot-setup}
\end{minipage}
\end{figure}

\subsubsection{Domain-Specific Reasoner}
\label{sec:domain-specific-reasoner}
To realize the cognitive functions required for interactive program synthesis, we contribute a novel reasoning system consisting of a Python executable (the control module) and a KnowRob plugin (the reasoner). The control module controls the program generation process, using the Prolog-based reasoners in KnowRob and the natural language interface (see Sec.~\ref{sec:natural-language-interface}) to gather information about the workflow, tools, workpieces, etc., and to make decisions about what actions to take. The reasoner is implemented as a set of Prolog predicates over classes, individuals, and relations in the knowledge base. Using these predicates, the control module can not only access explicit knowledge, but also derive new facts about existing knowledge. One example is the predicate \texttt{has\_tool(Task, Tool)}, which is \texttt{true} for all tool types that can be used for a given task.

\subsection{Natural Language Interface}
\label{sec:natural-language-interface}
At runtime, some information required for program synthesis, such as the exact material of the workpiece, cannot be inferred from the knowledge base and must be obtained by interacting with the user. For this purpose, we leverage an \ac{nlp} pipeline combining the Google Speech Recognition API \cite{noauthor_speech--text_nodate}, based on deep neural networks, and the spaCy \cite{honnibal_spacy_nodate} \ac{nlp} library:
\begin{enumerate}
    \item Conversion of the user's speech to text using the Google Speech Recognition API.
    \item Grammar-based chunking of the text into semantic units using spaCy.
    \item Semantic matching of chunks to the concept(s) to be grounded (e.g., ``mm'' to \texttt{Unit}, or ``fiberglass'' to \texttt{Material}), based on spaCy's word vector distance.
\end{enumerate}
Voice output (text-to-speech) is realized with the Windows voice APIs.

\subsection{Interactive Programming Workflow}
\label{sec:interactive-programming-workflow}
Fig. \ref{fig:meta-wizard-sequence-diagram} shows part of the interactive robot programming workflow. The complete process takes the following steps:
\begin{enumerate}
    \item \textbf{Meta-task grounding}: The user is prompted for the top-level task (e.g., \texttt{Sanding}). Suitable tools, possible workpieces and task parameters (such as the amount of material to be removed) are retrieved from the knowledge base. Missing knowledge is grounded by asking the user.
    \item \textbf{Task selection}: To take steps towards achieving the meta-task, the next (sub-)task is identified by querying the knowledge base for the next successor in the meta-tasks workflow definition, depending on the meta-wizard's current belief state. In some cases, this requires input from the user, e.g., when deciding whether to perform an additional pass over the surface.
    \item \textbf{Task execution}: The selected task (e.g., \texttt{Simulation}) is executed by calling a corresponding handler function in the control module. 
    \item \textbf{Iteration}: Iteration of steps 2) and 3) until a terminal step in the workflow is reached.
\end{enumerate}
When the information provided by the user is insufficient, e.g. due to misunderstood voice input, this information is discarded and the current step is repeated.

\section{Force-Controlled Surface Treatment}
\label{sec:force-controlled-surface-treatment} 
\subsection{Path Planning}
\label{sec:path-planning}
According to the position and orientation of the scanned surface, a set of intersection planes is adaptively defined. By slicing the point cloud uniformly with multiple parallel intersection planes, a set of cross-section contours is determined. This requires identifying all points whose distance from a  plane does not exceed a certain threshold. These points are then projected onto the corresponding plane and subsequently sorted, filtered, and interpolated to form the corresponding contours. As a continuous path is desired, a curve determination process enables generating sets of points that allow connecting the adjacent contours to form one meandering path (see Fig.~\ref{fig:path_planning}). The path planner is designed to operate on vertically projectively planar surfaces. We refer to \cite{iff2023} for more technical details and discussion of the planner.

\subsection{Simulation and Code Generation}
The generated robot program and tool path are simulated and visualized in a 3D environment. The 3D digital twin of the robot is updated by the meta-wizard during program generation to reflect, e.g., the surface finishing tool. Simulation both permits to validate the kinematic feasibility of the planned path and the collision-freeness of approach and depart motions, but also permits the user to ensure safe execution of the program. The program is then compiled to executable robot code and executed on the robot. For simulation, compilation, and execution, the robot programming software ArtiMinds RPS\footnote{\href{https://www.artiminds.com/}{ArtiMinds Robotics GmbH}, Karlsruhe, Germany} is used.

\subsection{Hybrid Force-Position Control}
At runtime, the planned path is executed by a hybrid force-position controller. The control law is a PID controller
\begin{align*}
u_\mathrm{wrench}(t) &= K_p e(t) + K_i\int_0^te(\tau)d\tau + K_d\frac{de(t)}{dt}~,\\
u(t) &= u_\mathrm{pose}(t) + u_\mathrm{wrench}(t)~.
\end{align*}
The control signal $u(t)$ is a 6-dimensional vector denoting a spatial offset (Cartesian position and orientation), which is the sum of separate wrench and pose components. The 6D wrench component $u_\mathrm{wrench}(t)$ is computed by a standard PID law, where the wrench error $e(t)$ is the distance between the currently measured end-effector wrench to the allowed wrench region (a 6D hypercube spanned by the wrenches allowed for the application). For sanding of fiberglass, this region collapses to a point, where the force $F_z$ is applied along the $z$-dimension, and zero along all other dimensions. The pose component $u_\mathrm{pose}(t)$ is the deviation of the current end-effector pose from the point on the planned path at time~$t$.

\section{Experiments}

The following comprehensive experimental evaluation of RoboGrind assesses its capabilities regarding robot-assisted surface treatment for remanufacturing, exemplified by wind turbine blades. The perception, planning, and control subsystems are individually assessed under laboratory conditions. Moreover, the overall system is evaluated holistically under real conditions. In all experiments, we utilize a UR10e robot\footnote{\href{https://www.universal-robots.com/}{Universal Robots} A/S, Odense, Denmark} with a disk sander\footnote{\href{https://onrobot.com/de}{OnRobot GmbH}, Soest, Germany} for safe human-robot collaboration. As robots by ABB, KUKA, FANUC or MOTOMAN are commonly used for polishing tasks \cite{zeng2023surface}, we additionally provide an assessment of the suitability of UR robots with respect to surface treatment.

\subsection{Laboratory Experiments}
\subsubsection{Perception}
\label{sec:lab_experiments_perception}
Three identical 500\,mm x 750\,mm pieces of the same uncoated fiberglass rotor blade with the same curvature are partitioned into four segments, resulting in a total of twelve concave segments with various surface defects (indentations and rough areas). The segments are scanned, followed by filling of the defects with fiberglass putty, scanned again, and finally all the segments are scanned a third time after sanding. The result is 36 point clouds, 12 without defects and 24 with 96 detectable defects. The detected, undetected and falsely detected defects are marked on the 24 point clouds with detectable defects.

\begin{table}
     \centering
     \caption{Evaluation of the path planning algorithm w.r.t. to its ability of approximating the surface to be sanded. Metrics are computed for each of the 12 point clouds independently. Averages and standard deviations are reported.}
		\begin{tabular}{ccc}
            \toprule
			RMSE / mm & MAE / mm & MAX / mm \\
			\midrule
			0.372 {\scriptsize($\sigma=0.030$)} & 0.348 {\scriptsize(0.034)} & 1.046 {\scriptsize(0,330)} \\
		\bottomrule
		\end{tabular}
	\label{table:evaluation_path_planning}
\end{table}

\subsubsection{Planning}

Precise path planning is characterized by how well the path aligns with the surface it approximates. To measure this alignment, we determine the distance from each point on the path to the closest point within the surface point cloud obtained after being primed for subsequent sanding. Using these distances, we estimate the accuracy of the approximation by computing the metrics \ac{rmse}, \ac{mae}, and \ac{max}.  It is worth mentioning that the scores computed are directly correlated with the resolution of the surface point cloud, i.e., higher resolution leads to lower scores in general and vice versa, requiring normalization for ideal comparability. We report the raw scores without normalization as all point clouds have nearly the same amount of total points ($\mu = 83,501.75$, $\sigma = 2,288.78$), which introduces a small but defensible bias in favor of reporting values with an associated dimension (mm).

\subsubsection{Control}

For each of the 36 scans, the surface is sanded at least once using the proposed simulation, planning, and control subsystems. Robot end-effector trajectories, comprising end-effector forces, are measured and analyzed. Each surface segment is sanded with different parameters, varying the contact pressure $F_z$, angle of attack $\alpha$ (in $^{\circ}$) and number of passes. The contact pressures used in the $z$ direction were 5\,N and 10\,N, the angles of attack between the disc and the workpiece were 2\,° and 5\,°, and the process was repeated for 5 or 10 passes. A total of 120 executions have been performed. To quantify the precision of the force controller, we compute the \ac{mae} and \ac{max} error of the measured end-effector forces along the $z$-axis with respect to the contact force setpoint $F_z$. In addition, we compute the rise time of the controller, defined as the time it takes for the measured end-effector force to reach 90\,\% of $F_z$.

\subsection{Real-World Experiments}
To evaluate RoboGrind under realistic conditions, we deploy it on-site at a company specialized in robot-based surface treatment\footnote{\href{https://www.shl.ag/de}{SHL AG}, Böttingen, Germany}. In contrast to the laboratory experiments, the experiments are conducted on the complete blade and not on a cut-out section. First, a scan of the surface and two sanding passes are performed (rough and fine sanding). Then fiberglass putty is applied, the surface is scanned again and two additional passes are performed. The main aim of the experiment is to determine to what extent RoboGrind can provide user assistance for robotic surface treatment. The experiment setup is shown in Fig.~\ref{fig:shl-robot-setup}.

\section{Results}
\subsection{Laboratory Experiments}
\subsubsection{Perception}
Evaluation of laboratory experiments regarding the perception and damage detection demonstrated that perception is robust in over 71\,\% of the 24 samples and 96 included damages. 69 defects were correctly detected, 27 were missed and 7 marked spots were false positives. Remaining problems are holes in point clouds caused by failed measurements and false positive defect detections. Failed measurements result from darker colored or dirty surfaces that are not detectable by the laser scanner. False positives occur in sections where the depth of the defects is low because of the dynamic thresholds of the algorithms.

\subsubsection{Planning} The planner performs robustly on all 12 point clouds (see Table \ref{table:evaluation_path_planning} and Fig.~\ref{fig:path_planning}). Scores ranging from 0.331\,mm to 0.425\,mm for \ac{rmse} and 0.307\,mm to 0.420\,mm for \ac{mae} indicate a consistent performance, although some point clouds do contain void areas, leading to higher \ac{max} scores.

\subsubsection{Control}
\begin{table}
    \centering
    \caption{Measured force control metrics for the three tested contact force threshold values $F_z$. \ac{max} is computed after $F_z$ has first been reached.}
    \label{tab:forcecontrol_results}
    \input{table_forcecontrol_results}    
\end{table}

\begin{figure}
\centering
\includegraphics[width=\linewidth]{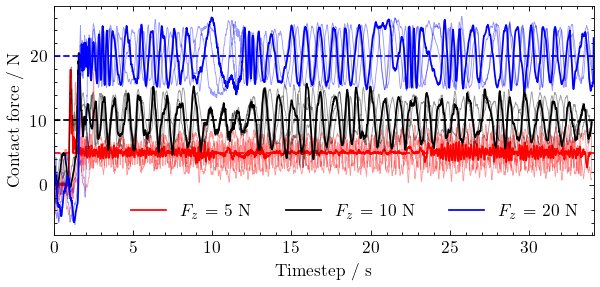}
\caption{Four randomly selected, exemplary executions for each of the three tested contact forces. One execution per group has been highlighted. Smaller force setpoints lead to better controller behavior.}
\label{fig:controller_behavior}
\end{figure}

Out of the 120 sanding attempts, 69\,\% were successful. The remainder failed due to exceeding force-torque limits of the UR10e, which occurred because of the natural vibration of the workpiece. For the 83 successful executions, the computed metrics are shown in Table~\ref{tab:forcecontrol_results}. The \ac{mae} increases with the force setpoint, but is limited to 2.1\,N, indicating sufficiently precise control for fiberglass sanding. The rise time remained below 1 second for $F_z=5$. Overall, we found that a low force setpoint of 5\,N achieves the best quantitative controller behavior and resulting surface quality. Exemplary force trajectories for each force setpoint are plotted in Fig.~\ref{fig:controller_behavior}.

In the laboratory experiments, we found that the sanding parameters shown in Table \ref{table:evaluation_sanding} deliver the best results. Additionally to these parameters the rotational speed and the traverse speed were set to 6,000\,rpm and 20\,m/s, respectively. The surface roughness of the rotor blade before the process is about 90\,$\mu$m. We compare the roughness after filling $R_\mathrm{a1}$ to roughness after sanding $R_\mathrm{a2}$, which is close to the initial value.

\subsection{Real-World Experiments}
In the field experiment, we found that all components of RoboGrind performed as well in an industrial setting as in the laboratory. The 3D scanning and path planning algorithms performed robustly under realistic, uncontrolled light conditions, both before and after putty has been applied. For the path planner, we report errors close to the values observed in the laboratory (0.378\,mm, 0.357\,mm and 0.994\,mm for \ac{rmse}, \ac{mae} and \ac{max}). After putty has been applied, small regions with strong noise regarding the scanned surface led to noisy point normals and therefore a slight, but sudden change in the orientation of the sander. This can be avoided by applying a global low-pass filter on the point cloud before path planning. The interactive meta-wizard understood user inputs well despite loud background noise. The control behavior is influenced by the much larger dimensions of the workpiece (see Fig.~\ref{fig:shl-robot-setup}) and the resulting strongly differing natural frequencies. A sufficient surface quality for the application was achieved. We leave the optimization of controller parameters for future work. Overall, the UR10e robot proved suitable for sanding tasks.

\setlength{\tabcolsep}{4pt}

\begin{table}
    \centering
    \caption{Evaluation of the surface quality in laboratory experiments. Measured surface roughness for four parameter sets.}
    \label{table:evaluation_sanding}
    \begin{tabular}{ccccc}
        \toprule
        \multicolumn{3}{c}{Parameters} & \multicolumn{2}{c}{Metrics} \\
        \cmidrule(lr){1-3} \cmidrule(lr){4-5} 
        $F_{z}$ / N & $\alpha$ / $^{\circ}$ & Passes & $R_\mathrm{a1}$ / $\mu$m & $R_\mathrm{a2}$ / $\mu$m \\
        \midrule
        10 & 5 & 5 & 199.57 & 122.44  \\
        10 & 2 & 10 & 169.45 & 104.25  \\
        5 & 5 & 5 & 123.46 & 162.02 \\
        5 & 2 & 10 & 225.01 & 119.82 \\

        \bottomrule
    \end{tabular}
\end{table}

\section{Conclusions}
In this work, we introduced RoboGrind, an \ac{ai}-assisted robotic system for automating surface treatment tasks. 

We contribute the first quantitative evaluation of robotic sanding of fiberglass with a collaborative robot. Moreover, we provide a qualitative evaluation on a real-world use case.

Our findings indicate that RoboGrind is able to largely automate the process from perception to program execution for surface treatment. The system components achieve a level of reliability and competence suitable for real-world use.

However, some limitations remain. The meta-wizard is currently limited to tasks within its existing knowledge base---future work will explore methods for learning new tasks from demonstration or observation. Applications beyond surface treatment, such as peg-in-hole assembly, also remain open directions for future work.






\section*{Acknowledgment}

This work was supported by the state of Baden-Württemberg in the project RoboGrind under grant BW1\_0079/01, the DFG CRC EASE (CRC \#1320) and the EU project euROBIN (grant 101070596).


\bibliographystyle{IEEEtran}
\bibliography{references_dhbw,references_amr, references_iff}

\end{document}

%% file: table_forcecontrol_results.tex
\setlength{\tabcolsep}{4pt}
\begin{tabular}{ccccc}\toprule
\multicolumn{1}{c}{Parameters} & \multicolumn{3}{c}{Metrics} & \\
\cmidrule(lr){1-1}\cmidrule(lr){2-4}
$F_z$ / N & MAE / N & MAX / N & Rise time & Trials\\\midrule
5 & 1.47 & 9.74 & 0.96 & 40\\
10 & 1.84 & 7.17 & 1.75 & 26\\
20 & 2.10 & 5.52 & 1.35 & 17\\
\bottomrule
\end{tabular}